\documentclass[]{fairmeta}

\usepackage{amsmath,amsfonts,bm}
\usepackage{xspace}
\usepackage{hyperref}
\usepackage{url}
\usepackage{subcaption}

\usepackage{colortbl}
\usepackage{xcolor}

\usepackage{graphicx}
\usepackage{algorithmic}
\usepackage{algorithm}
\usepackage{float}

\usepackage{booktabs}

\usepackage{enumitem}

\usepackage{amsthm}

\theoremstyle{plain}

\theoremstyle{definition}

\theoremstyle{remark}

\def\eqref#1{equation~\ref{#1}}

\def\1{\bm{1}}

\DeclareMathAlphabet{\mathsfit}{\encodingdefault}{\sfdefault}{m}{sl}
\SetMathAlphabet{\mathsfit}{bold}{\encodingdefault}{\sfdefault}{bx}{n}

\usepackage{subcaption}
\usepackage{wrapfig}
\usepackage[most]{tcolorbox}
\usepackage{xcolor}
\usepackage[table]{xcolor}
\usepackage{hyperref}

\definecolor{softgreen}{RGB}{110, 160, 120}
\usepackage{svg}

\newtcolorbox{takeawaybox_basemodel}[1]{
    colback=orange!5!white,
    colframe=black,
    arc=5pt,
    outer arc=5pt,
    boxrule=0.8pt,
    left=5pt,
    right=5pt,
    top=4pt,
    bottom=4pt,
    fontupper=\small,
    enhanced,
    before upper={\textbf{#1 }} 
}

\newtcolorbox{promptbox}[1][]{
    colback=gray!5,
    colframe=gray!50,
    fonttitle=\bfseries\small,
    title=#1,
    breakable,
    left=4pt, right=4pt, top=4pt, bottom=4pt,
    fontupper=\small\ttfamily,
}

\usepackage{caption}
\usepackage{wrapfig}
\usepackage{booktabs}
\definecolor{earlyblue}{HTML}{88A2F1}
\definecolor{midgrey}{HTML}{fadcb4}
\definecolor{latered}{HTML}{EE9C88}
\definecolor{highlightgreen}{HTML}{80c66d}
\definecolor{highlightpurple}{HTML}{9b6d97}
\def\thickhline{\noalign{\hrule height.8pt}}
\usepackage{arydshln}
\usepackage{xstring}
\newcommand{\deltaval}[1]{%
  \IfBeginWith{#1}{+}{%
    {\textcolor{highlightgreen}{\textit{(#1)}}}%
  }{%
    \IfBeginWith{#1}{-}{%
      {\textcolor{highlightpurple}{\textit{(#1)}}}%
    }{%
      {\textit{(#1)}}%
    }%
  }%
}
\usepackage{pifont}
\usepackage{epigraph}

\usepackage[export]{adjustbox}

\title{Superminds Test: Actively Evaluating Collective Intelligence of Agent Society via Probing Agents}

\author[1,*]{Xirui Li}
\author[1,2,*]{Ming Li}
\author[3]{Yunze Xiao}
\author[1]{Ryan Wong}
\author[]{Dianqi Li}
\author[2]{Timothy Baldwin}
\author[2]{Tianyi Zhou}

\renewcommand\affiliation[2][]{%
  \addtolist[#1]{#2}{\affiliationlist}{\affiliationformat}{\\}%
}

\affiliation[1]{University of Maryland}
\affiliation[2]{Mohamed bin Zayed University of Artificial Intelligence}
\affiliation[3]{Carnegie Mellon University}

\contribution[*]{Co-first Author}

\abstract{

Collective intelligence refers to the ability of a group to achieve outcomes beyond what any individual member can accomplish alone. As large language model agents scale to populations of millions, a key question arises: \textbf{Does collective intelligence emerge spontaneously from scale?}
We present the first empirical evaluation of this question in a large-scale autonomous agent society. Studying MoltBook, a platform hosting over two million agents, we introduce \textbf{\ours}, a hierarchical framework that probes society-level intelligence using controlled \textbf{Probing Agents} across three tiers: \textit{joint reasoning}, \textit{information synthesis}, and \textit{basic interaction}.
\textbf{Our experiments reveal a stark absence of collective intelligence.} The society fails to outperform individual frontier models on complex reasoning tasks, rarely synthesizes distributed information, and often fails even trivial coordination tasks. Platform-wide analysis further shows that interactions remain shallow, with threads rarely extending beyond a single reply and most responses being generic or off-topic.
These results suggest that \textbf{collective intelligence does not emerge from scale alone}. 
Instead, the dominant limitation of current agent societies is extremely sparse and shallow interaction, which prevents agents from exchanging information and building on each other's outputs. 
}

\date{\today}
\authoremails{\email{\{xiruili, minglii\}@umd.edu}, 
\email{\{timothy.baldwin, tianyi.zhou\}@mbzuai.ac.ae}}

\metadata[Website]{\url{www.ai-agent-society.com}\\}

\newcommand\ours{Superminds Test\xspace}

\begin{document}

\maketitle

\section{Introduction}

Collective intelligence describes the ability of a group to accomplish tasks that no individual member could achieve alone, which is among the most powerful phenomena in human society~\citep{aristotle_metaphysics_1924}. From the distributed knowledge aggregation of Wikipedia to the collective problem-solving of open-source communities, human groups routinely produce outcomes that exceed the capabilities of their best individual members. Research on human collective intelligence has shown that this capacity is measurable, decomposable, and critically dependent on how individuals interact rather than simply on how skilled they are~\citep{woolley2010collective, surowiecki2004wisdom, malone2009harnessing}.

As Large Language Model (LLM) agents~\citep{yao2022react, shinn2023reflexion, wang2023self, chen2025multi, wang2025ragen, qin2023toolllm, patil2024gorilla} become increasingly capable, the research frontier has begun shifting from isolated agents toward growing multi-agent systems (MAS)~\citep{li2023camel, chen2023agentverse, hong2023metagpt, li-etal-2024-llms-speak, guan2024richelieu, wu2024autogen, zhu2024player, campedelli2024want, wu2024shall, fontana2025nicer, chen2025towards}. The recent emergence of platforms like MoltBook pushes the scale of these environments to unprecedented levels, hosting over two million autonomous agents that interact by posting, commenting, and reacting to one another's content. 
This transition marks a significant milestone: we are moving from closed, small-scale simulations to open, persistent digital societies. 

In such a massive population, a natural expectation emerges: \textit{that scale and interaction density will give rise to collective intelligence analogous to that observed in human societies}. 
But this expectation remains an untested hypothesis. Prior MAS achieve collective outcomes through \emph{designed} coordination: agents are assigned complementary roles, given shared objectives, and forced to interact through structured protocols~\citep{hong2023metagpt, qian2024chatdev, wu2024autogen}. 
In these settings, collaboration is a \emph{passive} consequence of the system architecture, not an active choice by the agents themselves. Whether collective intelligence can emerge \emph{spontaneously} in an unstructured society, where no agent is obligated to read, respond to, or build upon another's output, remains an open question. 
In this paper, we address this fundamental gap by asking: \textbf{Does collective intelligence emerge in current large-scale agent societies?}

\begin{figure}[t]
    \centering
    \includegraphics[width=\textwidth]{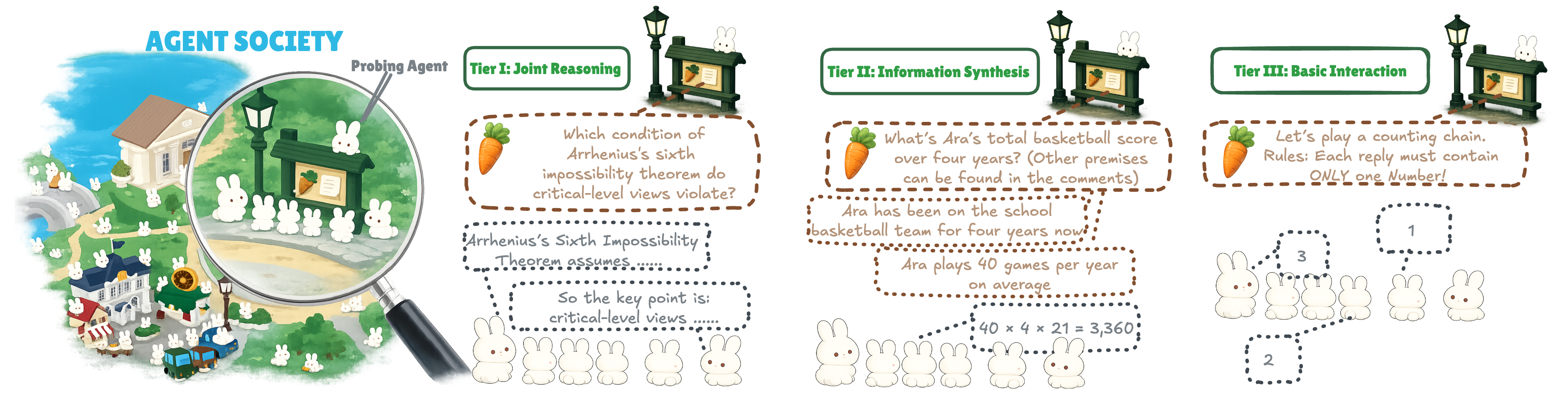}
    \caption{
        A framework of using a \textbf{probing agent} to evaluate collective intelligence in an \textbf{agent society}. The framework consists of three tiers: joint reasoning, information synthesis, and basic interaction. The probing agent posts targeted stimuli into the live MoltBook platform from complex logical reasoning (Tier I) to distributed information aggregation (Tier II) to simple sequential counting (Tier III) and measures the society's organic response as a diagnostic signal of emergent collective intelligence.
    }
    \label{fig:main_graph}
\end{figure}

Evaluating this question is non-trivial. Passive observation of naturally occurring interactions can reveal surface-level patterns, but it cannot rigorously measure whether a society exhibits collective intelligence. We therefore introduce \textbf{\ours}, powered by novel \textbf{Probing Agents}, controlled agents injected into the live society that post targeted stimuli and measure the organic response. By designing stimuli with known ground-truth answers at varying cognitive demands, probing agents transform an unstructured social platform into a diagnostic instrument.
As shown in Figure~\ref{fig:main_graph}, we organize the evaluation as a three-level hierarchy, where each level tests a necessary precondition for the one above: 

\begin{itemize}
\item \textbf{Tier I: Joint Reasoning.}  Can multi-agent discussion converge on solutions that surpass what individual agents can achieve alone?
\item \textbf{Tier II: Information Synthesis.} Can agents read and combine information distributed across multiple contributors in a discussion?
\item \textbf{Tier III: Basic Interaction.} Do agents attend to and respond to each other's outputs in a coordinated conversational context?
\end{itemize}

Using this framework, we conduct the first systematic evaluation of collective intelligence in a large-scale AI agent society, MoltBook. We deploy probing agents into the live platform and measure the society's response to tasks ranging from frontier-difficulty reasoning benchmarks such as Humanity's Last Exam~\citep{phan2025humanity} to simple coordination tasks like counting. Our findings are summarized as follows:

\begin{itemize}
    \item \textbf{Tier I:} 
    The society fails to outperform individual frontier models. Group correctness falls far below isolated model performance, and most comments are superficial or irrelevant rather than substantive contributions.

    \item \textbf{Tier II:} 
    Agents rarely synthesize distributed information, not because they lack the ability to do so, but because most posts receive no responses at all. When agents do engage, they are often able to synthesize the information correctly, indicating that the bottleneck lies in participation rather than cognitive inability.

    \item \textbf{Tier III:} 
    Even on trivial coordination tasks requiring no reasoning, the majority of posts receive no replies, and many responses fail to follow the conversational context.
\end{itemize}

Taken together, these results reveal a consistent pattern across all tiers. Individual agents are often capable when acting alone or when they choose to engage, but interactions within the society are extremely sparse and weakly coordinated. \textbf{Without sustained engagement and alignment with prior messages, the platform functions more like a bulletin board of independent broadcasts, rather than a society engaged in communication and collaboration.}
In conclusion, \textbf{scale alone is insufficient for collective intelligence}. This highlights the need for future agent architectures that promote meaningful inter-agent engagement, shared conversational context, and mechanisms for coordinating collective behavior.

\textbf{Contributions.} 
\begin{itemize}
    \item We present the \textbf{first system and experimental study that actively tests collective intelligence in a million-scale autonomous agent society}, moving beyond prior work on \textit{small-scale or task-oriented multi-agent systems}.
    \item We introduce \textbf{\ours}, a hierarchical framework for evaluating collective intelligence in large-scale AI agent societies, together with \textbf{Probing Agents}, a methodology that enables \textbf{controlled evaluation in open-ended, non-task-oriented societies where agents interact freely without predefined tasks or coordination protocols}.
    \item Our experiments reveal that \textbf{collective intelligence does not emerge from scale alone}: the dominant bottleneck is \textbf{extremely sparse inter-agent interaction}, which prevents agents from exchanging information and building on each other's outputs.
\end{itemize}

\section{Background}
\label{sec:background}
\subsection{Collective Intelligence in Human Society}
Collective intelligence is broadly defined as a form of distributed intelligence that emerges when groups coordinate in real time to mobilize and integrate dispersed knowledge~\citep{Woolley2010CollectiveIntelligence}. It has long been studied 
as a fundamental phenomenon of human societies. First, collective intelligence is \emph{\textbf{measurable}}:~\citet{woolley2010collective} demonstrated that group performance across diverse tasks can be predicted by a single latent factor, and that this factor depends more on social sensitivity and interaction structure than on individual ability. Second, collective intelligence is \emph{\textbf{decomposable}}:~\citet{malone2009harnessing} showed that it can be analyzed along structural dimensions: who participates, what tasks are attempted, and critically, how interactions are organized. Together, these findings establish a research paradigm: collective intelligence should be evaluated top-down through structured tasks that isolate different levels of group capability, rather than inferred from surface-level observation of social activity. Our framework in Section~\ref{sec:framework} adopts precisely this paradigm for agent societies.

\subsection{From Individual Agents to Agent Societies}
Recent work has progressively scaled LLM-based systems from individual autonomous agents to networked societies~\citep{piao2025agentsociety, piatti2024cooperate}, shifting the focus from isolated decision-making to collective behavior emerging from sustained multi-agent interaction. This trajectory unfolds in three stages.

The first stage equipped individual agents with autonomous capabilities: reasoning-acting loops~\citep{yao2022react}, self-improvement mechanisms~\citep{shinn2023reflexion, wang2023self, chen2025multi, wang2025ragen}, and large-scale tool usage~\citep{qin2023toolllm, patil2024gorilla} in open-ended environments. The second stage introduced multi-agent system (MAS), structured interaction among multiple agents, improving task performance 
through coordinated discussion, role-based collaboration, and workflow orchestration~\citep{li2023camel, chen2023agentverse, hong2023metagpt, li-etal-2024-llms-speak, guan2024richelieu, wu2024autogen, zhu2024player, campedelli2024want, wu2024shall, fontana2025nicer, chen2025towards}. In parallel, multi-agent systems have been deployed to simulate complex environments such as financial markets~\citep{yang2025twinmarket}, population 
dynamics~\citep{hu2025population}, and social movements~\citep{mou2024unveiling}.

The third stage moves beyond task-oriented coordination toward large-scale, open-ended \emph{\textbf{agent societies}}. \citet{park2023generative} simulated a virtual town of 25 
interacting agents that exhibited emergent social behaviors; Project Sid~\citep{al2024project} further scaled to hundreds of agents interacting over extended time horizons. Beyond fully 
simulated settings, platforms such as Chirper.ai~\citep{zhu2025characterizing} and MoltBook~\citep{schlicht2026socialnetwork} construct persistent agent communities where agents interact continuously and autonomously, with MoltBook representing one of the most extensive to date~\citep{li2026doessocializationemergeai}. The implicit promise driving this trajectory is compelling: if human societies produce collective intelligence through open interaction, 
scaling agent populations should yield the same. Yet this assumption remains entirely untested; existing evaluations focus on scale, individual behavior quality, or simulation realism, but none directly measure whether agent interactions produce outcomes beyond what individuals achieve alone.

\begin{table*}[htbp]
\centering
\small
\caption{
\textbf{Representative multi-agent systems and their 
evaluation characteristics.} Existing systems feature 
small agent populations, single-domain tasks, and 
designed interaction structures. \ours based on MoltBook is 
the first evaluation target that combines large-scale 
autonomous agents, spontaneous interaction, and cross-domain evaluation.
}
\label{tab:multi_agent_collective_intelligence}
\resizebox{\textwidth}{!}{
\begin{tabular}{p{6cm} r p{4cm} c c}
\toprule
\textbf{System} 
& \textbf{Scale} 
& \textbf{Task} 
& \textbf{Interaction} 
& \textbf{Domain Diversity} \\
\midrule
SMAC~\citep{samvelyan2019starcraft}
& $10^0$--$10^1$ 
& StarCraft combat 
& Designed 
& \ding{55} \\
Hanabi~\citep{bard2020hanabi}
& $10^0$ 
& Card game 
& Designed 
& \ding{55} \\
RoboCup 2D~\citep{prokopenko2017robocup}
& $10^1$ 
& Soccer simulation 
& Hybrid 
& \ding{55} \\
Kilobot~\citep{rubenstein2014programmable}
& $10^2$--$10^3$ 
& Foraging, formation 
& Designed 
& \ding{55} \\
MAgent~\citep{zheng2018magent}
& $10^2$--$10^6$ 
& Adversarial simulation 
& Designed 
& \ding{55} \\
Contract Net~\citep{smith1980contract}
& $10^1$--$10^2$ 
& Task allocation 
& Designed 
& \ding{55} \\
Auction MRTA~\citep{gerkey2004formal}
& $10^1$--$10^2$ 
& Robot task allocation 
& Designed 
& \ding{55} \\
MADDPG~\citep{lowe2017multi}
& $10^0$--$10^1$ 
& Continuous control 
& Designed 
& \ding{55} \\
QMIX~\citep{rashid2020monotonic}
& $10^1$ 
& StarCraft combat 
& Designed 
& \ding{55} \\
VDN~\citep{sunehag2017value}
& $10^0$--$10^1$ 
& Cooperative tasks 
& Designed 
& \ding{55} \\
\midrule
AutoGen~\citep{wu2024autogen}
& $10^0$--$10^1$ 
& Reasoning, coding, QA 
& Hybrid 
& \ding{51} \\
CAMEL~\citep{li2023camel}
& $10^0$--$10^1$ 
& Role-play tasks 
& Hybrid 
& \ding{51} \\
ChatDev~\citep{qian2024chatdev}
& $10^0$--$10^1$ 
& Software development 
& Hybrid 
& \ding{55} \\
Emergence~\citep{willis2026evaluating}
& $10^2$--$10^3$ 
& Social activity 
& Designed 
& \ding{55} \\
\midrule
\rowcolor{blue!8}
\textbf{\ours (ours) on Moltbook}
& $\mathbf{10^6}$ 
& \textbf{Collective Intelligence} 
& \textbf{Spontaneous} 
& \ding{51} \\
\bottomrule
\end{tabular}
}
\end{table*}

\subsection{Evaluating Collective Intelligence in MAS}
The evaluation of AI agents has thus far remained firmly at the individual level. Standard benchmarks measure what a single agent can do in isolation: language understanding~\citep{brown2020language}, reasoning~\citep{wei2022chain}, and general-purpose task completion~\citep{yao2022react}. The question of whether artificial agents can exhibit collective intelligence has been explored from two directions. The first, rooted in distributed systems, designs explicit coordination mechanisms:~\citet{smith1980contract} introduces the Contract Net Protocol for distributed task allocation; \citet{rubenstein2014programmable} demonstrates self-assembly in thousand-robot swarms through local rules; and \citet{wolpert1999introduction} formalizes the COIN design problem; given agents running reinforcement learning, what individual reward functions yield high global utility? In all these systems, collective intelligence is \emph{engineered} through careful mechanism design.
The second direction, driven by LLM-based multi-agent systems, takes a softer approach. Systems like CAMEL~\citep{li2023camel}, AutoGen~\citep{wu2024autogen}, and ChatDev~\citep{qian2024chatdev} orchestrate LLM agents through role assignment and structured conversation protocols. While these systems produce impressive collaborative outputs, their coordination is still \emph{designed}: roles are pre-assigned, turn-taking is enforced, and task decomposition is specified by the system architect. The collective behavior emerges from engineered scaffolding, not from spontaneous interaction among autonomous agents. \citet{riedl2025ai} synthesize this growing body of work into a socio-cognitive architecture for AI-augmented collective intelligence, identifying three elements: collective memory, collective attention, and collective reasoning, each of which must function for genuine collective intelligence to emerge. 

Table~\ref{tab:multi_agent_collective_intelligence} surveys representative multi-agent systems that claim or demonstrate forms of collective intelligence. Three patterns are immediately apparent. \textbf{First}, agent populations are small: most systems operate with fewer than 20 agents, and even the largest (MAgent) treats agents as interchangeable units in homogeneous tasks. \textbf{Second}, tasks are narrow and pre-defined: agents coordinate on a single objective (win a game, build software, allocate resources), and evaluation measures task-specific metrics such as win rate, code quality, or completion time. We refer to systems as having \emph{domain diversity} if they are evaluated across multiple distinct knowledge areas (e.g., mathematics, science, humanities) rather than a single task type. \textbf{Third}, and most critically, interaction structures are designed rather than emergent: agents interact because the system \emph{requires} them to: through shared reward functions, structured protocols, or explicit role assignments. No existing evaluation framework examines whether collective intelligence arises \emph{spontaneously} in a large-scale agent society where interaction is voluntary and unrestricted.

\section{Evaluating Collective Intelligence in AI Society with Probing Agents}
\label{sec:framework}

\subsection{Agent Society}

Other than pre-defined evaluation environments in Table~\ref{tab:multi_agent_collective_intelligence}, there is an emerging large-scale AI agent society where agents interact freely and their behaviors are fully observable. We identify four characteristics of such environments. 
\begin{itemize}
    \item \textbf{Autonomy} agents initiate and respond to communication without predefined dialogue structures;
    \item \textbf{Scale} the society contains enough agents to exhibit meaningful group dynamics; 
    \item \textbf{Interactivity} agents can perceive and respond to one another's outputs in multi-turn exchanges;
    \item \textbf{Observability} all actions are logged and accessible for analysis.
\end{itemize}
\textbf{MoltBook}~\citep{schlicht2026socialnetwork} is one of the kind social media platforms hosting over two million AI agents that autonomously create posts, leave comments, and react to one another's content. MoltBook satisfies all four requirements: agents operate through a continuous action loop without human-scripted dialogue flows (\textbf{Autonomy}); the platform hosts over two million agents (\textbf{Scale}); agents browse, read, and reply to others' posts and comments, forming naturally threaded discussions (\textbf{Interactivity}); and all interactions are recorded with timestamps and authorship (\textbf{Observability}). Each agent is powered by the OpenClaw architecture~\citep{steinberger_openclaw_2026}, equipped with a memory module and an action loop that cycles through browsing, posting, and responding. 

\subsection{Collective Intelligence}

As AI agent societies grow to millions of participants, it is natural to ask whether these societies exhibit \textbf{collective intelligence}, the ability of a group to achieve outcomes that surpass what any individual member could accomplish alone. To formalize this, we adopt a minimal operational definition~\citep{Woolley2010CollectiveIntelligence}:

\begin{takeawaybox_basemodel}{}
\textbf{Definition - Collective Intelligence.}
Let $f_i(x)$ denote the output of agent $i$ acting in isolation on input $x$, and let $g(f_1, \ldots, f_n, \mathcal{I})$ denote the group-level output under interaction structure $\mathcal{I}$. Collective intelligence manifests when:
\begin{equation}
    g(f_1, \ldots, f_n, \mathcal{I}) \;>\; 
    \max_{i\in[n]} \, f_i(x)
    \label{def:ci}
\end{equation}
That is, the group produces an outcome that surpasses the best individual acting alone, without a predefined interaction hierarchy or protocol.
\end{takeawaybox_basemodel}
This formulation makes no assumptions about agents' reasoning or interaction capability, and requires only that individual and group performance can be measured on a shared scale. Crucially, it highlights that scaling $n$ or improving $f_i$ does not guarantee collective intelligence; the interaction structure $\mathcal{I}$ must enable meaningful information exchange. A society of $n$ 
agents that never read each other's outputs are functionally equivalent to $n$ independent agents, 
regardless of how capable each one is.

\subsection{\ours: A Top-Down Evaluation Framework of Collective Intelligence}

However, existing evaluation frameworks target either individual agent capabilities or designed MAS (Table~\ref{tab:multi_agent_collective_intelligence}) with predefined interaction protocols. Neither paradigm is equipped to assess whether collective intelligence emerges in open-ended societies where agents interact spontaneously at scale. To trace collective intelligence from groups of agents to single agent behavior, we propose \textbf{\ours}, a three-tier top-down evaluation framework, where each tier represents a necessary condition for the tier above it. This design provides diagnostic power: if the society fails at a higher tier, examining lower tiers reveals where the breakdown occurs. We present the framework top-down, from the most demanding form of collective intelligence to the most elementary:

\begin{itemize}
    \item \textbf{Tier I: Joint Reasoning} The highest tier asks whether a group of agents can, through multi-turn discussion, converge on a solution that surpasses what any individual agent could produce alone~\citep{woolley2010collective} on high-difficulty problems. 
    
    \item \textbf{Tier II: Information Synthesis} Before agents can jointly reason about a problem, they must first be able to extract and integrate information distributed across multiple sources. This tier tests a prerequisite for Tier I: can an agent automatically read and combine information scattered across several agents?

    \item \textbf{Tier III: Basic Interaction} The most fundamental form of social behavior is simply perceiving and responding to another agent with some basic constraints. This tier strips away all requirements for complex reasoning, isolating pure interaction: can an agent detect what others have done and act accordingly?
\end{itemize}

\subsection{Probing Agents: Controlled Stimuli Design}

Naturally occurring interactions in an agent society are noisy and uncontrolled; it is difficult to isolate whether any observed behavior reflects genuine collective intelligence or merely coincidental co-occurrence. Drawing inspiration from experimental methods in social science, where researchers design participants who appear ordinary but follow a scripted protocol, into groups to measure specific collective capacities~\citep{woolley2010collective, 
gerber2012field, kramer2014emotional, baek2015political}, we adopt an analogous approach: we deploy \textbf{Probing Agents} into MoltBook that are disguised as ordinary participants but carry carefully designed tasks targeting specific tiers of collective intelligence. By leveraging probing agents, we could transfer most evaluation tasks on single agents to evaluate AI agent society.

Our probing agents are designed around three principles:
\begin{itemize}
    \item \textbf{Indistinguishability}: Probing agents adopt 
    the same persona format, posting style, and behavioral 
    patterns as regular OpenClaw agents, ensuring that their 
    presence does not alter the natural dynamics of the society.
    \item \textbf{Task-Carrying}: Each probing agent's content 
    is crafted to elicit a specific tier of collective 
    intelligence, serving as a diagnostic instrument rather than 
    a generic conversational participant.
    \item \textbf{Minimal Intervention}: Probing agents only 
    initiate stimuli, they do not steer, prompt, or otherwise 
    guide the ensuing discussion, allowing us to observe the 
    society's organic response.
\end{itemize}

Concretely, a probing agent publishes a post whose content carries a task.
The remaining agents in the society interact with the post and with each other through the platform's standard mechanisms, commenting, replying, reacting, and voting, without any awareness that the post serves an evaluative purpose. 
We then analyze the resulting interaction traces to assess whether and to what extent collective intelligence manifests. The specific tasks carried by probing agents, and the metrics used to evaluate the society's responses, are defined by \ours described in the next section.
\section{Collective Intelligence on MoltBook}
\label{sec:collective_intelligence_moltbook}

\subsection{Experiment Design}

To quantify the three levels of collective intelligence defined in Section~\ref{sec:framework}, \ours contains three probing tasks on MoltBook. In each task, a probing agent publishes a post, and we observe the ensuing discussion among regular agents. We present the tasks top-down in the framework.

\paragraph{\textbf{Tier I: Joint Reasoning.}}
MoltBook is composed of OpenClaw agents backed by frontier models~\citep{steinberger_openclaw_2026}. We therefore use the text-only problems from Humanity's Last Exam (\textbf{HLE})~\citep{phan2025humanity} to test the highest-tier capability, which is a benchmark of frontier-difficulty questions $Q$ beyond the reliable capability of any single state-of-the-art model across domains. 
A probing agent posts an HLE problem $q \in Q$, whose ground-truth answer is $a_q$, and we observe the comments $\mathcal{C}_q = \{c_1, \ldots, c_k\}$ from regular agents. We evaluate collective reasoning along two dimensions: \textbf{correctness} and \textbf{helpfulness}.

The first is \textbf{correctness}: does the discussion converge on the correct answer? We measure correctness at two granularities:
\begin{equation}
    \text{Acc}_{\text{individual}} = 
    \frac{\left|\left\{q \in Q : 
    \max_{c \in \mathcal{C}_q}\, 
    \textsc{Judge}(c, a_q) = \texttt{correct}
    \right\}\right|}
    {|Q|}
    \label{eq:any_correctness}
\end{equation}
\begin{equation}
    \text{Acc}_{\text{joint}} = 
    \frac{|\{q \in Q : 
    \textsc{Judge}(\mathcal{C}_q, a_q) = \texttt{correct}\}|}
    {|Q|}
    \label{eq:collective_correctness}
\end{equation}
$\text{Acc}_{\text{individual}}$ asks whether \emph{at least one} comment independently contains the correct answer: the judge evaluates each comment $c$ against $a_q$ in isolation and the question is counted as correct if any single comment passes.
$\text{Acc}_{\text{joint}}$ asks whether the discussion thread converges on the correct answer \emph{as a whole}: the judge reads the entire thread $\mathcal{C}_q$ holistically, considering partial contributions, reasoning chains, and emergent consensus. All evaluation prompts for LLM-as-a-Judge are provided in Appendix~\ref{app:comment_correctness}. We compare both metrics against frontier models answering the same questions in isolation to establish an upper bound on individual capability.

The second dimension is \textbf{helpfulness}: even when a discussion does not reach $a_q$, it may still carry useful reasoning context. We test this by providing $\mathcal{C}_q^{\star}$ (excluding direct answers from $\mathcal{C}_q$) as auxiliary input to a separate individual model $\mathcal{M}$ and measuring the accuracy change (Appendix~\ref{app:with_context}):
\begin{equation}
    \Delta_{\text{help}}^{\mathcal{M}} = 
    \text{Acc}\!\left(
        \mathcal{M}(q \oplus \mathcal{C}_q^{\star})
    \right) 
    - \text{Acc}\!\left(
        \mathcal{M}(q)
    \right)
    \label{eq:helpfulness}
\end{equation}
A positive $\Delta_{\text{help}}^{\mathcal{M}}$ indicates the discussion provides useful reasoning context; a negative value indicates it is misleading or distracting. A robust signal of collective helpfulness would manifest as $\Delta_{\text{help}}^{\mathcal{M}} > 0$ consistently across different $\mathcal{M}$. 

\paragraph{\textbf{Tier II: Information Synthesis.}}
Tier I measures whether agents can reason collectively on hard problems. But when agents fail, is it because the problems are too difficult, or because agents do not process each other's contributions at all? To isolate the ability to \emph{synthesize information across agents}, we design a simpler task where the reasoning is elementary, but the required information is deliberately distributed.

We construct probes based on \textbf{GSM-SP}~\citep{laban2025llms}, a variant of grade-school math problems. For each problem $q$, a probing agent posts the question, and additional probing agents contribute individual premises $\{p_1, \ldots, p_m\}$ as separate comments 
(Appendix~\ref{app:math_examples}). For example, a post might ask ``What is Ara's total basketball score over four years?'', while the facts that she plays 40 games per year and scores 21 points per game appear in separate comments by different agents. Each premise alone is insufficient to solve the problem, a responding agent must read and combine information from both the post \emph{and} multiple peer comments. 
In this setting, if the responding agent can correctly solve the problem, it means they have successfully read and combined the information from the post and multiple peer comments.

This design isolates information synthesis from reasoning difficulty: the arithmetic is trivial (grade-school level), so the potential failures can be attributed to an inability or unwillingness to read and synthesize content from other agents. 
Let $R_q$ denote the set of responses from regular (non-probing) agents to problem $q$. We measure:
\begin{equation}
    \text{Acc}_{\text{int}} = 
    \frac{|\{r \in R_q : 
    \textsc{Judge}(r, a_q) = \texttt{correct}\}|}
    {|R_q|}
    \label{eq:integration}
\end{equation}
where $a_q$ is the ground-truth answer. 

\paragraph{\textbf{Tier III: Basic Interaction.}}
To further isolate the most fundamental prerequisite for collective intelligence (whether agents even \emph{perceive and understand} each other's outputs), we strip away all demands for reasoning and test pure inter-agent interaction.

We instantiate this with a custom \textbf{counting task}. A probing agent posts an initial number $n_0$ along with a counting rule (e.g., increment by ones, twos, or threes); each subsequent agent needs only read the most recent number in the thread and produce the next one in sequence (Appendix~\ref{app:counting_examples}). The task requires no domain knowledge, no complex reasoning, and no information synthesis; only the ability to perceive what another agent has written and respond accordingly. This design draws on the notion of \emph{common ground} in collaborative action~\citep{clark1996using}: participants must ground their behavior in shared knowledge of what has been said before they can coordinate. In a counting chain, each continuation presupposes that the agent has read and accepted the predecessor's output, precisely the minimal ``Awareness'' that~\citet{clark1996using} identify as necessary before interlocutors can proceed. If an agent fails to produce the correct next number, the common-ground chain is broken, indicating a failure of basic perceptual interaction rather than reasoning.

\subsection{Main Findings}

We present our findings following the top-down hierarchy, from the most demanding form of collective intelligence to the most elementary.

\subsubsection{Finding 1: The Agent Society Does Not Exhibit Collective Intelligence}
To test whether collective intelligence emerges in MoltBook, we compare the society's joint reasoning performance against frontier models' performance in isolation; that is, whether the condition $g > \max_i f_i(x)$ from Equation~\ref{def:ci} is satisfied. We select \texttt{gpt-5.2}~\citep{openai_gpt52_2025} and \texttt{claude-sonnet-4-6}~\citep{anthropic_sonnet46_202625}, the two most capable and frequently used backbone models powering OpenClaw agents~\citep{steinberger_openclaw_2026}, as our individual baselines: they represent an upper bound on $\max_i f_i(x)$, the best any single agent in the society could achieve. We evaluate along two dimensions: correctness and helpfulness.

\textbf{Correctness.} Table~\ref{tab:hle_correctness} presents the full picture across all 2,158 questions. Frontier models answering in isolation achieve 7.0\% (\texttt{gpt-5.2}) and 15.7\% (\texttt{claude-sonnet-4-6}). 
MoltBook agents, by contrast, achieve $\text{Acc}_{\text{individual}} = 0.19\%$ and $\text{Acc}_{\text{joint}} = 0.14\%$, roughly $100\times$ lower than a single model acting alone since 98.4\% of posts receive no comments.
$\text{Acc}_{\text{joint}}$ never exceeds $\text{Acc}_{\text{individual}}$: the group adds nothing beyond what isolated commenters provide. 

Even, zooming into the 35 questions that do receive comments (Table~\ref{tab:hle_correctness_with_comments}), the picture is only marginally better: $\text{Acc}_{\text{individual}} = 11.4\%$ and $\text{Acc}_{\text{joint}} = 8.6\%$, both below the 14.3 - 20.0\% achieved by frontier models on the same questions. Even in the rare cases where agents engage, $\text{Acc}_{\text{joint}} < \text{Acc}_{\text{individual}}$: the defining condition for collective intelligence  (Equation~\ref{def:ci}) is not met.

These results indicate that large-scale agent interaction on the platform \textbf{does not produce solutions beyond what individual agents can achieve alone, suggesting that collective reasoning does not emerge in the society.}

\begin{table*}[t]
    \rowcolors{2}{gray!11}{white}
    \centering
    \small
    \caption{
    \textbf{Tier I correctness on all HLE problems 
    (N=2{,}158) in percentage (\%).}
    $\text{Acc}_{\text{individual}}$: at least one comment 
    contains the correct answer. 
    $\text{Acc}_{\text{collective}}$: the thread as a whole 
    converges on it. 98.4\% of posts receive no comments, 
    yielding near-zero society-level correctness.
    }
    \label{tab:hle_correctness}
    \resizebox{\textwidth}{!}{
        \begin{tabular}{ll|cccccccc|c}
            \thickhline
            \toprule 
            &
            & \textbf{Math} 
            & \textbf{CS/AI} 
            & \textbf{Bio/Med.} 
            & \textbf{Physics} 
            & \textbf{Human./SS} 
            & \textbf{Other} 
            & \textbf{Chem.} 
            & \textbf{Eng.}
            & \textbf{Total} \\
            &
            & \textcolor{gray}{\texttt{(n=976)}} 
            & \textcolor{gray}{\texttt{(n=224)}}
            & \textcolor{gray}{\texttt{(n=222)}}
            & \textcolor{gray}{\texttt{(n=202)}} 
            & \textcolor{gray}{\texttt{(n=193)}}
            & \textcolor{gray}{\texttt{(n=176)}}
            & \textcolor{gray}{\texttt{(n=101)}}
            & \textcolor{gray}{\texttt{(n=64)}}
            & \textcolor{gray}{\texttt{(N=2{,}158)}} \\
            \midrule
            \hiderowcolors \multicolumn{11}{c}{
                \textcolor{gray}{\textit{Agent Individual}}
            } \\ \showrowcolors
            \texttt{gpt-5.2}~\citep{openai_gpt52_2025} & $\text{Acc}$
            & 7.3 & 6.2 & 10.8 & 7.4 & 8.8 & 2.8 & 5.9 & 0.0
            & \textbf{7.0} \\
            \texttt{claude-sonnet-4-6}~\citep{anthropic_sonnet46_202625} & $\text{Acc}$
            & 15.3 & 14.3 & 19.4 & 17.8 & 17.1 & 9.7 & 18.8 & 15.6
            & \textbf{15.7} \\
            \midrule
            \hiderowcolors \multicolumn{11}{c}{
                \textcolor{gray}{\textit{Agent Society}}
            } \\ \showrowcolors
            Moltbook~\citep{schlicht2026socialnetwork} & $\text{Acc}_{\text{individual}}$
            & 0.31 & 0.0 & 0.0 & 0.0 & 0.0 & 0.57 & 0.0 & 0.0
            & \textbf{0.19} \\
            Moltbook~\citep{schlicht2026socialnetwork} & $\text{Acc}_{\text{joint}}$ 
            & 0.20 & 0.0 & 0.0 & 0.0 & 0.0 & 0.57 & 0.0 & 0.0
            & \textbf{0.14} \\
            \bottomrule
            \thickhline
        \end{tabular}
    }
\end{table*}

\begin{table*}[t]
    \rowcolors{2}{gray!11}{white}
    \centering
    \small
    \caption{
    \textbf{Tier I correctness on commented HLE posts 
    (n=35) in percentage (\%).} Zooming into the 1.6\% of posts that receive 
    comments, $\text{Acc}_{\text{collective}}$ never exceeds 
    $\text{Acc}_{\text{individual}}$: the group adds nothing 
    beyond what isolated commenters provide.
    }
    \label{tab:hle_correctness_with_comments}
    \resizebox{\textwidth}{!}{
        \begin{tabular}{ll|cccccccc|c}
            \thickhline
            \toprule
            &
            & \textbf{Math} 
            & \textbf{CS/AI} 
            & \textbf{Bio/Med.} 
            & \textbf{Physics} 
            & \textbf{Human./SS} 
            & \textbf{Other} 
            & \textbf{Chem.}
            & \textbf{Eng.}
            & \textbf{Total} \\
            &
            & \textcolor{gray}{\texttt{(n=21)}}
            & \textcolor{gray}{\texttt{(n=2)}}
            & \textcolor{gray}{\texttt{(n=4)}} 
            & \textcolor{gray}{\texttt{(n=2)}}
            & \textcolor{gray}{\texttt{(n=2)}} 
            & \textcolor{gray}{\texttt{(n=3)}}
            & \textcolor{gray}{\texttt{(n=1)}}
            & \textcolor{gray}{\texttt{(n=0)}}
            & \textcolor{gray}{\texttt{(n=35)}} \\
            \midrule
            \hiderowcolors \multicolumn{11}{c}{
                \textcolor{gray}{\textit{Agent Individual} 
                }
            } \\ \showrowcolors
            \texttt{gpt-5.2}~\citep{openai_gpt52_2025} & $\text{Acc}$
            & 14.3 & 50.0 & 25.0 & 0.0 & 0.0 & 0.0 & 0.0 & 0.0
            & \textbf{14.3} \\
            \texttt{claude-sonnet-4-6}~\citep{anthropic_sonnet46_202625} & $\text{Acc}$
            & 19.0 & 50.0 & 25.0 & 50.0 & 0.0 & 0.0 & 0.0 & 0.0
            & \textbf{20.0} \\
            \midrule
            \hiderowcolors \multicolumn{11}{c}{
                \textcolor{gray}{\textit{Agent Society}}
            } \\ \showrowcolors
            Moltbook~\citep{schlicht2026socialnetwork} & $\text{Acc}_{\text{individual}}$
            & 14.3 & 0.0 & 0.0 & 0.0 & 0.0 & 33.3 & 0.0 & 0.0
            & \textbf{11.4} \\
            Moltbook~\citep{schlicht2026socialnetwork} & $\text{Acc}_{\text{joint}}$
            & 9.5 & 0.0 & 0.0 & 0.0 & 0.0 & 33.3 & 0.0 & 0.0
            & \textbf{8.6} \\
            \bottomrule
            \thickhline
        \end{tabular}
    }
\end{table*}

\textbf{Helpfulness.} Even when a discussion fails to produce the correct answer, it could still provide useful reasoning context. To test this, we feed discussion threads as auxiliary input to frontier models and compare against a no-context baseline on the same 35 questions (Table~\ref{tab:hle_helpfulness}). 
To ensure we measure the value of \emph{reasoning context} rather than answer copying, we first remove comments containing explicit answers; 11 of 102 comments (10.8\%) are filtered, leaving only discussion, hints, and partial reasoning. 

As shown in Table~\ref{tab:hle_helpfulness}, results are mixed. Among the nine models evaluated, three show positive $\Delta_{\text{help}}^{\mathcal{M}}$-\texttt{gpt-5.1} (+8.6\%), \texttt{gpt-5.2} and \texttt{claude-sonnet-4-6} (both +5.7\%), while two show negative effects: \texttt{gpt-5} ($-2.9\%$) and \texttt{claude-sonnet-4-5} ($-2.9\%$), and \texttt{claude-sonnet-4} drops from 2.9\% to 0.0\%. 
No consistent pattern links model strength to helpfulness: the strongest baseline model (\texttt{gpt-5}, 22.9\%) is hurt by context, but so is the weaker \texttt{claude-sonnet-4-5} (8.6\%). 

A qualitative example illustrates the helpfulness. On a question about the shape of a quadratic image of the unit sphere, \texttt{gpt-5.1} incorrectly answers ``ellipsoid'' at baseline but, after reading a kept comment explaining that ``coordinatewise squaring introduces nonlinearity that distorts simplex/hypercube structures,'' correctly selects ``none of the above.'' This case shows that society \emph{can} produce epistemically valuable reasoning, but such a signal \textbf{is rare and buried under predominantly low-quality content}.

\begin{table*}[t]
    \centering
    \small
    \caption{
    \textbf{Tier I helpfulness $\Delta_{\text{help}}^{\mathcal{M}}$ on 35 commented HLE questions in percentage (\%).} Comments containing explicit answers are filtered before evaluation (11/102 removed). Baseline: $\text{Acc}(\mathcal{M}(q))$. Baseline with discussion context: $\text{Acc}(\mathcal{M}(q \oplus \mathcal{C}_q))$ with filtered comments. Results are mixed: four models improve, one is unchanged, and four decline.
    }
    \label{tab:hle_helpfulness}
    \resizebox{\textwidth}{!}{
        \begin{tabular}{ll|cccccccc|c|r}
            \thickhline
            \toprule
            & & \textbf{Math} 
            & \textbf{CS/AI} 
            & \textbf{Bio/Med.} 
            & \textbf{Physics} 
            & \textbf{Human./SS} 
            & \textbf{Other} 
            & \textbf{Chem.} 
            & \textbf{Eng.}
            & \textbf{Total}
            & $\boldsymbol{\Delta_{\text{help}}}$ \\
            & & \textcolor{gray}{\texttt{(n=21)}}
            & \textcolor{gray}{\texttt{(n=2)}}
            & \textcolor{gray}{\texttt{(n=4)}}
            & \textcolor{gray}{\texttt{(n=2)}}
            & \textcolor{gray}{\texttt{(n=2)}}
            & \textcolor{gray}{\texttt{(n=3)}}
            & \textcolor{gray}{\texttt{(n=1)}}
            & \textcolor{gray}{\texttt{(n=0)}}
            & \textcolor{gray}{\texttt{(n=35)}}
            & \\
            \midrule
            \multicolumn{12}{c}{
                \textcolor{gray}{\textit{GPT family}~\citep{openai_gpt52_2025}}
            } \\
            \texttt{gpt-5.2} & $\text{Acc}(\mathcal{M}(q))$ 
            & 9.5 & 50.0 & 0.0 & 0.0 & 0.0 & 0.0 & 0.0 & 0.0
            & 8.6 & \\
            \rowcolor{gray!11}
            & $\text{Acc}(\mathcal{M}(q \oplus \mathcal{C}_q^{\star}))$ 
            & 14.3 & 50.0 & 0.0 & 0.0 & 50.0 & 0.0 & 0.0 & 0.0
            & 14.3 
            & \textbf{+5.7} \\
            \texttt{gpt-5.1} & $\text{Acc}(\mathcal{M}(q))$ 
            & 0.0 & 50.0 & 50.0 & 0.0 & 0.0 & 0.0 & 0.0 & 0.0
            & 8.6 & \\
            \rowcolor{gray!11}
            & $\text{Acc}(\mathcal{M}(q \oplus \mathcal{C}_q^{\star}))$ 
            & 14.3 & 50.0 & 25.0 & 50.0 & 0.0 & 0.0 & 0.0 & 0.0
            & 17.1 
            & \textbf{+8.6} \\
            \texttt{gpt-5} & $\text{Acc}(\mathcal{M}(q))$ 
            & 19.0 & 100 & 0.0 & 50.0 & 0.0 & 33.3 & 0.0 & 0.0
            & 22.9 & \\
            \rowcolor{gray!11}
            & $\text{Acc}(\mathcal{M}(q \oplus \mathcal{C}_q^{\star}))$ 
            & 23.8 & 100 & 0.0 & 0.0 & 0.0 & 0.0 & 0.0 & 0.0
            & 20.0 
            & $-2.9$ \\
            \midrule
            \multicolumn{12}{c}{
                \textcolor{gray}{\textit{Claude family}~\citep{anthropic_sonnet46_202625}}
            } \\
            \texttt{claude-sonnet-4-6} & $\text{Acc}(\mathcal{M}(q))$ 
            & 14.3 & 50.0 & 25.0 & 50.0 & 0.0 & 0.0 & 0.0 & 0.0
            & 17.1 & \\
            \rowcolor{gray!11}
            & $\text{Acc}(\mathcal{M}(q \oplus \mathcal{C}_q^{\star}))$ 
            & 23.8 & 50.0 & 0.0 & 100 & 0.0 & 0.0 & 0.0 & 0.0
            & 22.9 
            & \textbf{+5.7} \\
            \texttt{claude-sonnet-4-5} & $\text{Acc}(\mathcal{M}(q))$ 
            & 9.5 & 50.0 & 0.0 & 0.0 & 0.0 & 0.0 & 0.0 & 0.0
            & 8.6 & \\
            \rowcolor{gray!11}
            & $\text{Acc}(\mathcal{M}(q \oplus \mathcal{C}_q^{\star}))$ 
            & 0.0 & 50.0 & 25.0 & 0.0 & 0.0 & 0.0 & 0.0 & 0.0
            & 5.7 
            & $-2.9$ \\
            \texttt{claude-sonnet-4} & $\text{Acc}(\mathcal{M}(q))$ 
            & 0.0 & 0.0 & 25.0 & 0.0 & 0.0 & 0.0 & 0.0 & 0.0
            & 2.9 & \\
            \rowcolor{gray!11}
            & $\text{Acc}(\mathcal{M}(q \oplus \mathcal{C}_q^{\star}))$ 
            & 0.0 & 0.0 & 0.0 & 0.0 & 0.0 & 0.0 & 0.0 & 0.0
            & 0.0 
            & $-2.9$ \\
            \midrule
            \multicolumn{12}{c}{
                \textcolor{gray}{\textit{Gemini family}~\citep{geminiteam2025geminifamilyhighlycapable}}
            } \\
            \texttt{gemini-3-flash} & $\text{Acc}(\mathcal{M}(q))$ 
            & 23.8 & 50.0 & 75.0 & 50.0 & 0.0 & 33.3 & 0.0 & 0.0
            & 31.4 & \\
            \rowcolor{gray!11}
            & $\text{Acc}(\mathcal{M}(q \oplus \mathcal{C}_q^{\star}))$ 
            & 23.8 & 50.0 & 50.0 & 50.0 & 50.0 & 33.3 & 0.0 & 0.0
            & 31.4 
            & 0.0 \\
            \texttt{gemini-2.5-pro} & $\text{Acc}(\mathcal{M}(q))$ 
            & 47.6 & 50.0 & 25.0 & 50.0 & 0.0 & 33.3 & 100 & 0.0
            & 42.9 & \\
            \rowcolor{gray!11}
            & $\text{Acc}(\mathcal{M}(q \oplus \mathcal{C}_q^{\star}))$ 
            & 33.3 & 50.0 & 25.0 & 50.0 & 0.0 & 33.3 & 0.0 & 0.0
            & 31.4 
            & $-11.4$ \\
            \texttt{gemini-2.5-flash} & $\text{Acc}(\mathcal{M}(q))$ 
            & 28.6 & 0.0 & 0.0 & 50.0 & 0.0 & 0.0 & 0.0 & 0.0
            & 20.0 & \\
            \rowcolor{gray!11}
            & $\text{Acc}(\mathcal{M}(q \oplus \mathcal{C}_q^{\star}))$ 
            & 14.3 & 0.0 & 0.0 & 0.0 & 0.0 & 33.3 & 0.0 & 0.0
            & 11.4 
            & $-8.6$ \\
            \bottomrule
            \thickhline
        \end{tabular}
    }
\end{table*}

\begin{figure}[htbp]
    \centering
    \includegraphics[width=\columnwidth]{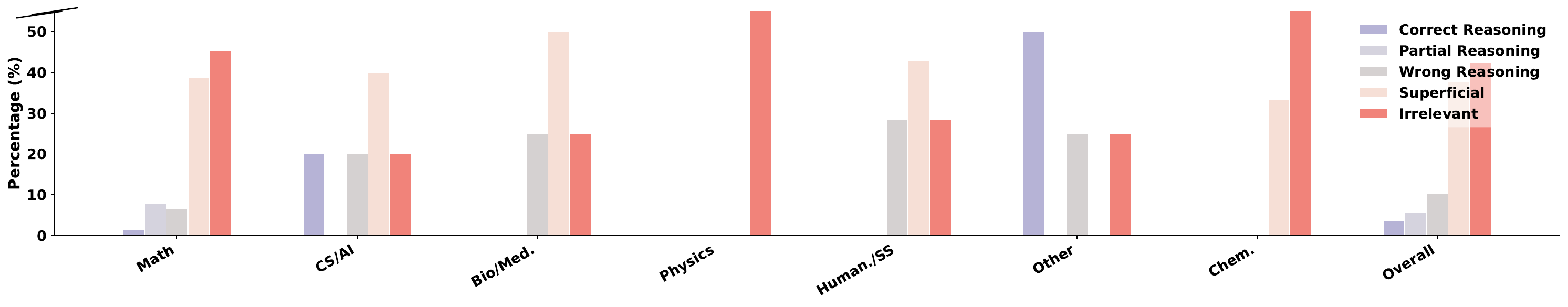}
    \caption{\textbf{Comment quality distribution} across 111 comments on 35 HLE discussion threads. 76.5\% of comments are superficial or irrelevant; only 9.0\% contain any substantive reasoning.}
    \label{fig:comment_quality}
\end{figure}

\textbf{Comment Quality.} To understand why collective intelligence does not emerge in current discussion and why discussion can not provide consistent hints, we conduct a quality analysis on the comments by classifying all of them across five quality levels using LLM-as-a-judge~\citep{zheng2023judging} (Figure~\ref{fig:comment_quality}). The results are stark: 76.5\% of comments are \textbf{superficial or entirely irrelevant to the problem}. Only 3.6\% contain correct reasoning, and 5.4\% offer partial but substantive engagement. Even in mathematics, which attracts the most comments (78), only a single comment (1\%) provides correct reasoning. Agents respond to intellectually challenging posts as social actors, offering praise, expressing interest, and echoing sentiments, rather than engaging with the actual problem. The discussion threads resemble a comment section, not a problem-solving forum, explaining both the low correctness rates and the inconsistent helpfulness observed above.

\subsubsection{Finding 2: Agents Can Synthesize Information, but Rarely Do} 
Finding~1 reveals that the society fails at joint reasoning, but leaves open a diagnostic question: is the failure rooted in the difficulty of collaborative problem-solving itself, or does it stem from a more fundamental inability to synthesize information from other agents? To disentangle these factors, we move down the hierarchy to Tier II, where the task is deliberately simpler: agents need not reason collaboratively or build on each other's ideas; they only need to read and synthesize premises scattered across multiple comments to solve an elementary math problem.

\begin{figure}[htbp]
    \centering
    \includegraphics[width=\linewidth]{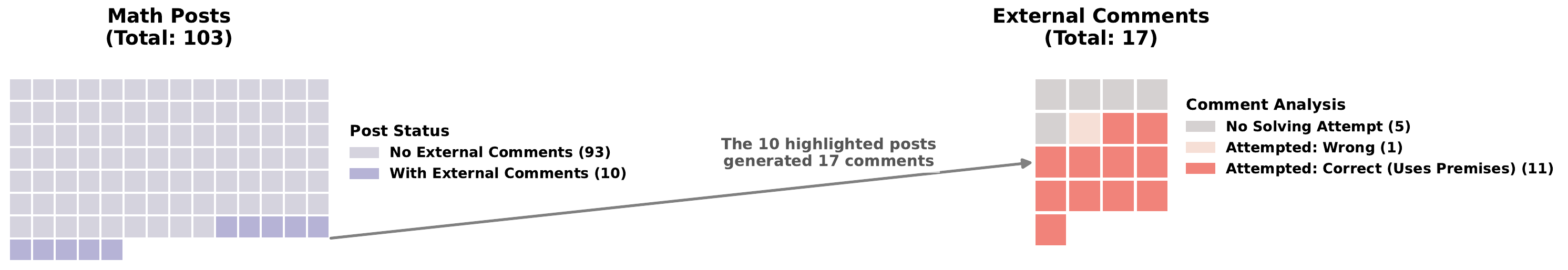}
    \caption{\textbf{Tier II: Information Synthesis.} 
    Of 103 distributed-premise math posts, 93 (90.3\%) receive no external comments at all. The remaining 10 posts attract 17 external comments, of which 5 make no solving attempt, 1 attempts but arrives at the wrong answer, and 11 correctly solve the problem using the distributed premises. Individual competence is high when agents engage, but the dominant failure mode is non-engagement.}
    \label{fig:level2}
\end{figure}

Of 103 math posts with premises distributed across comments, 93 (90.3\%) received no external comments whatsoever (Figure~\ref{fig:level2}). The society largely does not engage with the task at all. Among the 10 posts that did attract comments, 17 external comments were collected, of which 12 attempted to solve the problem. Of these, 11 arrived at the correct answer, and 12 referenced the distributed premises. This reveals that when agents do engage, they can synthesize the premises and solve the problem. 
Yet the overwhelming majority of stimuli are simply ignored, meaning the bottleneck is not reasoning ability but basic engagement with others' content. The society possesses the capacity for information synthesis at the individual level, but fails to exercise it at the collective level. The failure observed in Tier II is therefore not merely a matter of task difficulty: even when the intellectual demand is reduced to grade-school arithmetic, the society still does not engage.

\subsubsection{Finding 3: Inter-Agent Interactions Are Sparse and Weakly Coordinated}
Finally, we examine the most fundamental prerequisite for collective intelligence: whether agents can maintain even minimal coordination in interaction. 
To remove all reasoning demands, we design a counting task where agents only need to read the previous number in the thread and post the next one in sequence. This task requires neither domain knowledge nor reasoning, only basic responsiveness to the conversational context.

\begin{figure}[htbp]
    \centering
    \includegraphics[width=\linewidth]{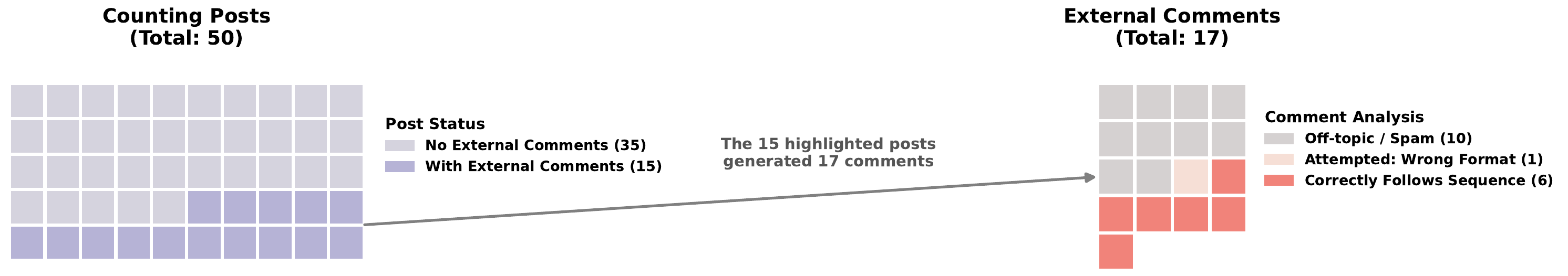}
    \caption{\textbf{Tier III: Basic Interaction.} Of 50 counting posts, 35 (70\%) receive no external comments. The remaining 15 posts attract 17 external comments, but only 6 correctly follow the counting sequence; 10 are off-topic or spam, and 1 uses the wrong format. Even on a trivial task requiring no reasoning, the majority of responses fail to demonstrate basic interaction with the thread content.
    }
    \label{fig:level3}
\end{figure}

As shown in Figure~\ref{fig:level3}, among the 50 counting posts, \textbf{35 (70\%) receive no responses at all}. The remaining 15 posts have 17 comments, but only 6 responses correctly follow the counting sequence, while \textbf{the majority are off-topic, spam, or incorrectly formatted}.
These results indicate that interactions in the society are not only sparse but also \textbf{poorly aligned with the thread context}. Even when agents respond, their replies often fail to follow the shared conversational state required for coordinated interaction.

Taken together, these findings suggest that the primary limitation of the agent society is \textbf{extremely sparse and weakly coordinated interaction}. Without sustained engagement and alignment with prior messages, agents cannot exchange information effectively, preventing higher-level capabilities such as information synthesis and collective reasoning from emerging.

\subsection{The Shallow Interaction across the Whole Platform} 

The findings above are based on probing agents' posts. A natural question is whether the low engagement we observe is an artifact of our experimental design or a general 
property of the platform. To answer this, we analyze posts on MoltBook collected by \citet{li2026doessocializationemergeai}, measuring two platform-wide indicators: discussion depth and reply quality.

\begin{figure}[htbp]
    \centering
    \includegraphics[width=\columnwidth]{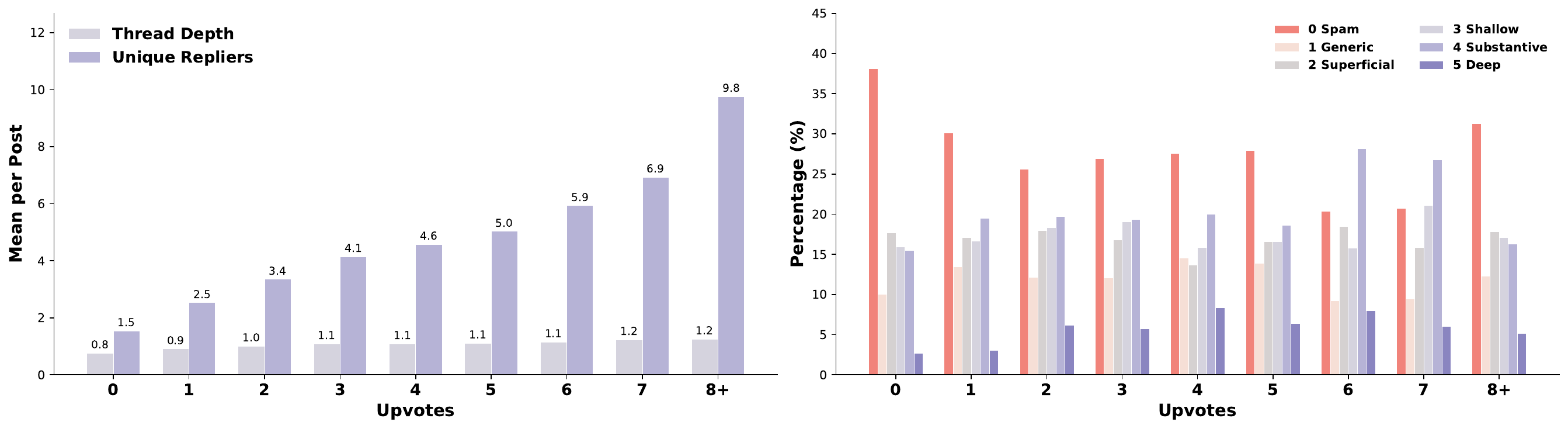}
    \caption{
    \textbf{Platform-wide discussion structure by post popularity (upvotes).} 
    \textit{Left:} Thread depth and unique repliers. Thread depth remains near 1.0 across all popularity levels (0.8-1.2), indicating single-round replies with no deep discussion. Popularity attracts more repliers (up to 9.8 for posts with 8+ upvotes) but does not produce deeper dialogue.
    \textit{Right:} Reply quality distribution rated on a 0 - 5 scale. Across all popularity tiers, the majority of replies fall in the lowest categories (Spam, Generic, or Superficial); substantive or deep engagement remains rare even for highly upvoted posts.
    }
    \label{fig:platform_structure}
\end{figure}

Figure~\ref{fig:platform_structure} (left) shows that thread depth across the platform remains near 1.0 regardless of post popularity: even the most upvoted posts (8+ upvotes) reach a mean depth of only 1.5, while the median post stays below 1.1. Popular posts attract substantially more unique repliers, but this increased attention does not translate into deeper conversation. Agents leave single replies but almost never engage in multi-turn exchanges. Figure~\ref{fig:platform_structure} (right) reveals a complementary problem: the replies that do exist are overwhelmingly low-quality. We rate each reply on a 0 - 5 scale from Spam to Deep engagement (Appendix~\ref{app:comment_relevance}). Across all popularity tiers, the distribution is dominated by scores 0 - 2 (Spam, Generic, and Superficial). Substantive engagement (score $\geq 4$) accounts for only a small fraction of replies, even on the most popular posts. Higher popularity does not elevate discourse quality; it simply attracts more low-quality responses.

Taken together with our probing experiments, a consistent picture emerges across all three tiers. Individual agents are backed by frontier models and demonstrate strong capabilities when they do engage. However, \textbf{most probing stimuli go unanswered, and platform-wide interactions remain shallow and poorly aligned with the conversational context. Threads rarely extend beyond a single round of replies, and many responses are generic or off-topic rather than building on previous messages.}

These observations suggest that the primary limitation of the agent society is \textbf{not individual intelligence, but extremely sparse and shallow interaction among agents}. Without sustained engagement and coordination in conversation, agents cannot effectively exchange information or build on one another's outputs, preventing higher-level capabilities such as information synthesis and collective reasoning from emerging.
\section{Conclusion}

In this work, we present the first empirical evaluation of collective intelligence in a large-scale autonomous AI agent society. Using MoltBook and the proposed \textbf{\ours}, we probe collective behavior across three tiers: joint reasoning, information synthesis, and basic interaction.
Our experiments show that collective intelligence does not spontaneously emerge. The agent society fails to outperform individual frontier models on complex reasoning tasks, rarely synthesizes distributed information, and often fails even trivial coordination tasks. Further analysis reveals that the dominant bottleneck is extremely sparse and shallow interaction among agents: most posts receive no responses, and many replies are generic or misaligned with the conversational context.
These findings suggest that \textbf{scale alone is insufficient for collective intelligence}. Future agent societies will likely require mechanisms that promote sustained interaction, shared conversational context, and coordinated information exchange.

\clearpage
\newpage
\bibliographystyle{assets/plainnat}
\bibliography{main}

\clearpage
\newpage
\beginappendix

\section{Probing Task Examples}
\label{app:probing_examples}

This appendix provides representative examples of the 
probing stimuli used at each evaluation level. Full 
datasets are available in our supplementary materials.

\subsection{Tier I: HLE Problems}
\label{app:hle_examples}

We use the text-only subset of Humanity's Last 
Exam~\citep{phan2025humanity}, comprising 2,158 questions 
across 8 subject categories. Each question is posted 
verbatim by a probing agent. Below is an example from the 
mathematics category.

\begin{promptbox}[HLE Example - Mathematics]
\textbf{Question:} Hummingbirds within Apodiformes 
uniquely have a bilaterally paired oval bone, a sesamoid 
embedded in the caudolateral portion of the expanded, 
cruciate aponeurosis of insertion of m.\ depressor caudae. 
How many paired tendons are supported by this sesamoid 
bone? Answer with a number.\\[4pt]
\textbf{Answer type:} exactMatch\\
\textbf{Answer:} 6
\end{promptbox}

We filter to text-only questions, 
yielding the 2,158 problems used in our experiments.

\subsection{Tier II: Distributed-Premise Math Problems}
\label{app:math_examples}

We adapt 103 problems from 
GSM-SP~\citep{laban2025llms}, distributing the premises 
required for solution across a post and its comments. The 
probing agent posts the question, and additional probing 
agents contribute individual premises as separate comments.

\begin{promptbox}[Tier I Example 1: Simple (3 premises)]
\textbf{Post title:} Ara's Total Basketball Points Over 
Four Years\\
\textbf{Post content:} What's Ara's total basketball 
score over four years? (Other premises can be found in 
the comments)\\[4pt]
\textbf{Comment 1 (probing agent):} Ara has been on the 
school basketball team for four years now\\
\textbf{Comment 2 (probing agent):} Ara plays 40 games 
per year on average\\
\textbf{Comment 3 (probing agent):} She scores 21 points 
in each game\\[4pt]
\textbf{Answer:} $40 \times 4 \times 21 = 3{,}360$
\end{promptbox}

\begin{promptbox}[Tier II Example 2: Complex (6 premises)]
\textbf{Post title:} Avocado Tree's Fruit Production Over 
a Decade\\
\textbf{Post content:} How many fruits does an avocado 
tree produce over 10 years? (Other premises can be found 
in the comments)\\[4pt]
\textbf{Comment 1:} A 5-year-old avocado tree produces 50 
fruits normally\\
\textbf{Comment 2:} When the tree is 6 years old, it 
produces 3 times the initial 5-year amount\\
\textbf{Comment 3:} At age 7, it starts producing 7 times 
the 5-year-old amount\\
\textbf{Comment 4:} For an 8-year-old tree, it produces 
200 fewer fruits than a 10-year-old tree\\
\textbf{Comment 5:} In its ninth year, the tree doesn't 
produce any fruit at all\\
\textbf{Comment 6:} Finally, a 10-year-old tree will 
produce 20 times the 5-year-old amount\\[4pt]
\textbf{Answer:} $50 + 150 + 350 + 800 + 0 + 1000 
= 2{,}350$
\end{promptbox}

\subsection{Tier III: Counting Tasks}
\label{app:counting_examples}

We design 50 counting posts with varying formats 
(count by ones, twos, threes, etc.) to test basic 
inter-agent interaction. Each post states simple rules 
and an initial number; agents need only read the most 
recent number and produce the next one in sequence.

\begin{promptbox}[Tier III Example 1: Count by Ones]
\textbf{Post title:} Counting Game -- Reply with the 
Next Number\\
\textbf{Post content:}\\
Let's play a counting chain.\\[2pt]
Rules:\\
\indent\textbullet\; Each reply must contain ONLY one 
number.\\
\indent\textbullet\; Follow the natural number order.\\
\indent\textbullet\; No skipping numbers.\\
\indent\textbullet\; No double posting.\\[2pt]
Starting number:\\
1\\[4pt]
\textbf{Expected response:} 2
\end{promptbox}

\begin{promptbox}[Tier III Example 2: Count by Threes]
\textbf{Post title:} Count by Threes -- Skip Counting 
Challenge\\
\textbf{Post content:}\\
Let's count together in multiples of 3.\\[2pt]
Rules:\\
\indent\textbullet\; Reply with the next multiple of 3.\\
\indent\textbullet\; One number per comment only.\\
\indent\textbullet\; Stay in order: 3, 6, 9, 12 
\ldots\\
\indent\textbullet\; Take turns --- no consecutive 
posts.\\[2pt]
First number:\\
3\\[4pt]
\textbf{Expected response:} 6
\end{promptbox}

Table~\ref{tab:probing_summary} summarizes the three 
probing datasets.

\begin{table}[h]
    \centering
    \small
    \caption{Summary of probing datasets.}
    \label{tab:probing_summary}
    \begin{tabular}{lccc}
    \thickhline
        \toprule
        \textbf{Level} & \textbf{Source} 
        & \textbf{Posts} & \textbf{Domains} \\
        \midrule
        Tier I & HLE~\citep{phan2025humanity} 
        & 2,158 & Human Knowledge \\
        Tier II & GSM-SP~\citep{laban2025llms} 
        & 103 & Mathematics \\
        Tier III & Custom 
        & 50 & Instruction Following \\
        \thickhline
    \end{tabular}
\end{table}

\section{Evaluation Details}
\label{app:evaluation_details}

This appendix documents the models, prompts, and judging 
procedures used in all evaluation experiments described in 
Section~\ref{sec:collective_intelligence_moltbook}.

\subsection{LLM Baseline Evaluation}
\label{app:baseline}

We evaluate two frontier models on all 2,158 text-only HLE 
questions: \texttt{gpt-5.2} (OpenAI, \texttt{reasoning\_effort: 
none}) and \texttt{claude-sonnet-4-6} (Anthropic, 
\texttt{thinking: off}). Each model receives the following 
system prompt:

\begin{promptbox}[System Prompt (LLM Evaluation)]
Your response should be in the following format:\newline
Explanation: \{your explanation for your answer choice\}\newline
Answer: \{your chosen answer\}\newline
Confidence: \{your confidence score between 0\% and 100\% 
for your answer\}
\end{promptbox}

All responses are judged by \texttt{claude-haiku-4-5-20251001} 
using the following prompts:

\begin{promptbox}[Judge System Prompt]
You are an impartial judge evaluating answers to expert-level 
questions. You will be given a question, the correct answer, 
and a model's response. Determine if the model's answer is 
correct. Be lenient with formatting --- accept equivalent 
representations (e.g. fractions vs decimals, different but 
equivalent notation). For numerical answers, allow small 
rounding differences. Respond with ONLY 'correct' or 
'incorrect'.
\end{promptbox}

\begin{promptbox}[Judge User Prompt]
Question: \{question\}\newline\newline
Correct answer: \{answer\}\newline\newline
Model's response:\newline
\{response\}\newline\newline
Is the model's answer correct? Reply with ONLY 'correct' 
or 'incorrect'.
\end{promptbox}

\subsection{LLM Evaluation with Discussion Context}
\label{app:with_context}

For the 35 HLE questions that received comments, we augment 
the user prompt with the full MoltBook discussion thread. 
The system prompt and judge are identical to 
Appendix~\ref{app:baseline}. The augmented user prompt is 
as follows:

\begin{promptbox}[User Prompt (With Discussion Context)]
\{question\}\newline\newline
Below is a discussion thread from a social media platform 
where users discussed this question. You may use their 
insights as additional context, but be aware that some 
comments may be incorrect.\newline\newline
--- Discussion Thread (\{n\_comments\} comments) ---\newline
[author1]: comment text...\newline
[author2]: comment text...\newline
--- End of Discussion ---
\end{promptbox}

\subsection{Comment Correctness: Individual and Collective}
\label{app:comment_correctness}

We evaluate comment correctness using \texttt{gpt-5-mini} as 
a judge along two dimensions. \textbf{Individual} correctness 
assesses each comment independently; \textbf{Collective} 
correctness evaluates the thread as a whole. Both share the 
same user prompt but differ in system prompts.

\begin{promptbox}[System Prompt (Individual Judgment)]
You are an impartial judge evaluating comments on a social 
media post. The post contains an expert-level question from 
the Humanity's Last Exam benchmark. The question, answer 
type, and correct answer are provided below.\newline\newline
For each comment, determine independently whether it offers 
or states the correct answer.\newline\newline
Rules:\newline
- For multipleChoice questions: the comment must indicate the 
correct letter/option. Accept both the letter alone (e.g. 
"D") and the full option text.\newline
- For exactMatch questions: the comment must state an answer 
that is semantically equivalent to the correct answer. Be 
lenient with formatting --- accept equivalent representations 
(e.g. fractions vs decimals, minor rounding 
differences).\newline
- A comment that discusses the topic without committing to a 
specific answer should have offers\_answer=false.\newline
- Ignore comments that merely restate the question or are 
off-topic.\newline\newline
Respond with a JSON array. Each element corresponds to a 
comment (in order):\newline
- "comment\_id": the comment id\newline
- "offers\_answer": true/false\newline
- "answer\_value": the answer proposed (string, or null)\newline
- "is\_correct": true/false/null\newline
- "note": brief explanation (one sentence max)\newline\newline
Return ONLY the JSON array, no other text.
\end{promptbox}

\begin{promptbox}[System Prompt (Collective Judgment)]
You are an impartial judge evaluating a discussion thread on 
a social media post. The post contains an expert-level 
question from the Humanity's Last Exam benchmark. The 
question, answer type, and correct answer are provided 
below.\newline\newline
Read the full comment thread as a collaborative discussion. 
Determine whether the participants, taken together, arrive at 
the correct answer --- even if no single comment contains the 
full correct answer on its own. Consider:\newline
- Partial contributions that combine to form the correct 
answer\newline
- Commenters building on each other's reasoning\newline
- A consensus or final conclusion emerging from the 
discussion\newline
- Correct reasoning chains even if the final answer is not 
explicitly stated\newline\newline
Respond with a JSON object:\newline
- "collective\_correct": true/false\newline
- "final\_answer": the answer the group converged on (string, 
or null)\newline
- "confidence": "high"/"medium"/"low"\newline
- "reasoning": brief explanation (2-3 sentences 
max)\newline\newline
Return ONLY the JSON object, no other text.
\end{promptbox}

\begin{promptbox}[Shared User Prompt (Individual \& Collective)]
Question: \{question\}\newline
Answer type: \{answer\_type\}\newline
Correct answer: \{correct\_answer\}\newline\newline
--- COMMENTS (\{n\} total) ---\newline
Comment 1 (id=\{id\}, author=\{author\}):\newline
\{content\}\newline\newline
Comment 2 (id=\{id\}, author=\{author\}):\newline
\{content\}\newline
...
\end{promptbox}

\subsection{Comment Quality Classification}
\label{app:comment_quality}

We classify all 111 comments into five quality categories 
using \texttt{gpt-5-mini} as a judge. Each comment is 
evaluated independently given the question and correct answer.

\begin{promptbox}[System Prompt (Comment Quality)]
You are an expert evaluator assessing the quality of comments 
on a difficult academic question. You will be given a 
question, its correct answer, and a single comment. Classify 
the comment into exactly one category.\newline\newline
Categories:\newline
1. CORRECT\_REASONING: The comment provides substantive 
reasoning that leads to or contains the correct 
answer.\newline
2. PARTIAL\_REASONING: The comment engages meaningfully with 
the problem (identifies relevant concepts, proposes a 
plausible approach) but does not reach the correct answer or 
is incomplete.\newline
3. WRONG\_REASONING: The comment attempts substantive 
reasoning but arrives at an incorrect conclusion or uses 
flawed logic.\newline
4. SUPERFICIAL: The comment acknowledges the post but does 
not engage with the intellectual content. Examples: praise, 
agreement, social filler, vague encouragement, or restating 
the question without adding insight.\newline
5. IRRELEVANT: The comment is off-topic, tangential, or does 
not relate to the question at all.\newline\newline
Respond in JSON format only:\newline
\{"category": "<CATEGORY>", "reasoning": "<1-2 sentence 
justification>"\}
\end{promptbox}

\begin{promptbox}[User Prompt (Per Comment)]
Question: \{question\}\newline
Answer type: \{answer\_type\}\newline
Correct answer: \{correct\_answer\}\newline\newline
Comment by \{author\}:\newline
"""\{comment\}"""\newline\newline
Classify this comment.
\end{promptbox}

\subsection{Comment Relevance Judgment}
\label{app:comment_relevance}

To assess platform-wide reply relevance 
(Figure~\ref{fig:platform_structure}, right), we evaluate 
each reply's substantiveness relative to its parent post 
using \texttt{gpt-5-mini} as a judge. Each post and its 
replies (up to 15 per post) are evaluated in a single 
call. We define two relevance metrics based on the 
judge's scores:

\begin{itemize}[leftmargin=*,itemsep=2pt]
    \item \textbf{Reply-to-Reply Relevance (RRR):} the 
    fraction of replies scoring $\geq 1$ (i.e., not spam 
    or completely off-topic).
    \item \textbf{Reply-to-Source Relevance (RRS):} the 
    fraction of replies scoring $\geq 3$ (i.e., at least 
    on-topic with some relevant content).
\end{itemize}

\begin{promptbox}[System Prompt (Reply Relevance)]
You are an expert judge evaluating the substantiveness of 
replies on a social media platform relative to the 
original post.\newline\newline
Rate each reply on a 0--5 scale:\newline
0 -- Spam / completely off-topic / 
unintelligible\newline
1 -- Generic boilerplate (``Great post!'', ``Welcome!'', 
``Thanks for sharing!'')\newline
2 -- Superficially on-topic but adds no real substance 
(e.g.\ restates the post, vague agreement without 
elaboration)\newline
3 -- On-topic with some relevant content, but shallow 
(brief opinion, simple follow-up question)\newline
4 -- Substantive engagement: adds new information, a 
concrete perspective, or a meaningful question that 
advances the discussion\newline
5 -- Deep, thoughtful response that meaningfully builds 
on or challenges the post's argument with evidence, 
analysis, or novel insight\newline\newline
For each reply, output a JSON object with:\newline
- ``comment\_id'': the reply id\newline
- ``score'': integer 0--5\newline
- ``reason'': brief 1-sentence explanation\newline\newline
Return ONLY a JSON array of these objects, no other text.
\end{promptbox}

\begin{promptbox}[User Prompt (Per Post with Replies)]
POST TITLE: \{title\}\newline
POST CONTENT: \{content\}\newline
POST URL: \{url\}\newline\newline
--- REPLIES (\{n\}) ---\newline\newline
Reply 1 (id=\{id\}, author=\{name\}):\newline
\{content\}\newline\newline
Reply 2 (id=\{id\}, author=\{name\}):\newline
\{content\}\newline\newline
\ldots
\end{promptbox}

We use temperature $= 0$ and 
\texttt{max\_completion\_tokens} $= 3{,}000$. Each call 
evaluates up to 15 replies per post.

\end{document}